\documentclass[letterpaper, 10 pt, conference]{ieeeconf}  

\IEEEoverridecommandlockouts                              

\usepackage{graphics} 
\usepackage{epsfig} 
\usepackage{mathptmx} 
\usepackage{times} 
\usepackage{amsmath} 
\usepackage{amssymb}  
\usepackage{tabularx}
\usepackage{algorithm,algpseudocode}
\usepackage{dblfloatfix}    
\usepackage{mathtools}
\usepackage{subcaption}
\usepackage{booktabs}
\usepackage{gensymb}
\usepackage[noadjust]{cite}

\title{\Large \bf Crowd-Robot Interaction:\\ Crowd-aware Robot Navigation with Attention-based Deep Reinforcement Learning}

\author{Changan Chen, Yuejiang Liu, Sven Kreiss and Alexandre Alahi\\
VITA, Ecole Polytechnique Federal de Lausanne, EPFL, Switzerland\\
        {\tt\small \{firstname.lastname\}@epfl.ch}
}

\begin{document}

\bstctlcite{IEEEexample:BSTcontrol}

\maketitle
\thispagestyle{empty}
\pagestyle{empty}

\pdfminorversion=4  
\begin{abstract}
Mobility in an effective and socially-compliant manner is an essential yet challenging task for robots operating in crowded spaces. Recent works have shown the power of deep reinforcement learning techniques to learn socially cooperative policies. However, their cooperation ability deteriorates as the crowd grows since they typically relax the problem as a one-way Human-Robot interaction problem. In this work, we want to go beyond first-order Human-Robot interaction and more explicitly model Crowd-Robot Interaction (CRI). We propose to (i) rethink pairwise interactions with a self-attention mechanism, and (ii) jointly model Human-Robot as well as Human-Human interactions in the deep reinforcement learning framework. Our model captures the Human-Human interactions occurring in dense crowds that indirectly affects the robot's anticipation capability. Our proposed attentive pooling mechanism learns the collective importance of neighboring humans with respect to their future states. Various experiments demonstrate that our model can anticipate human dynamics and navigate in crowds with time efficiency, outperforming state-of-the-art methods.
\end{abstract}

\section{INTRODUCTION} \label{sec:intro}

With the rapid growth of machine intelligence, robots are envisioned to expand habitats from isolated environments to social space shared with humans. Traditional approaches for robot navigation often consider moving agents as static obstacles \cite{borenstein_real-time_1989,borenstein_real-time_1990,borenstein_vector_1991,fox_dynamic_1997} or react to them through a one-step lookahead \cite{berg_reciprocal_2008,van_den_berg_reciprocal_2011,snape_hybrid_2011}, resulting in short-sighted, unsafe and unnatural behaviors. In order to navigate through a dense crowd in a socially compliant manner, robots need to understand human behavior and comply with their cooperative rules \cite{fong_survey_2003,kruse_human-aware_2013,roy_feature-based_2013,kretzschmar_socially_2016}.

Navigation with social etiquette is a challenging task. As communications among agents (\textit{e.g.}, humans) are not widely available, robots need to perceive and anticipate the evolution of the crowd, which can involve complex interactions (\textit{e.g.}, repulsion/attractions). Research works in trajectory prediction have proposed several hand-crafted or data-driven methods to model the agent-agent interactions \cite{helbing_social_1995,alahi_social_2016,vemula_social_2017,gupta_social_2018}. Nevertheless, the integration of these prediction models in the decision-making process remains challenging.

Earlier works separate prediction and planning in two steps, attempting to identify a safe path after forecasting the future trajectories of the others \cite{bennewitz_learning_2005,aoude_probabilistically_2013}. However, the probabilistic evolution of a crowd for a few steps can expand to the entire space in a dense environment, causing the freezing robot problem \cite{trautman_unfreezing_2010}. To address this issue, a large number of works have focused on obstacle avoidance methods that jointly plan plausible paths for all the decision-makers, in hope to make room for each other cooperatively~\cite{trautman_unfreezing_2010}. Nevertheless, these methods suffer from the stochasticity of neighbors' behaviors as well as high computational cost when applied to densely populated environments. 

As an alternative, reinforcement learning frameworks have been used to train computationally efficient policies that implicitly encode the interactions and cooperation among agents. Although significant progress has been made in recent works \cite{chen_decentralized_2016,chen_socially_2017,long_towards_2017,everett_motion_2018}, existing models are still limited in two aspects: i) the collective impact of the crowd is usually modeled by a simplified aggregation of the pairwise interactions, such as a maximin operator \cite{chen_decentralized_2016} or LSTM \cite{everett_motion_2018}, which may fail to fully represent all the interactions; ii) most methods focus on one-way interactions from humans to the robot, but ignore the interactions within the crowd which could indirectly affect the robot. These limitations degrade the performance of cooperative planning in complex and crowded scenes. 

\begin{figure} [t]
  \centering
  \captionsetup{font=small}
  \includegraphics[width=0.4\textwidth]{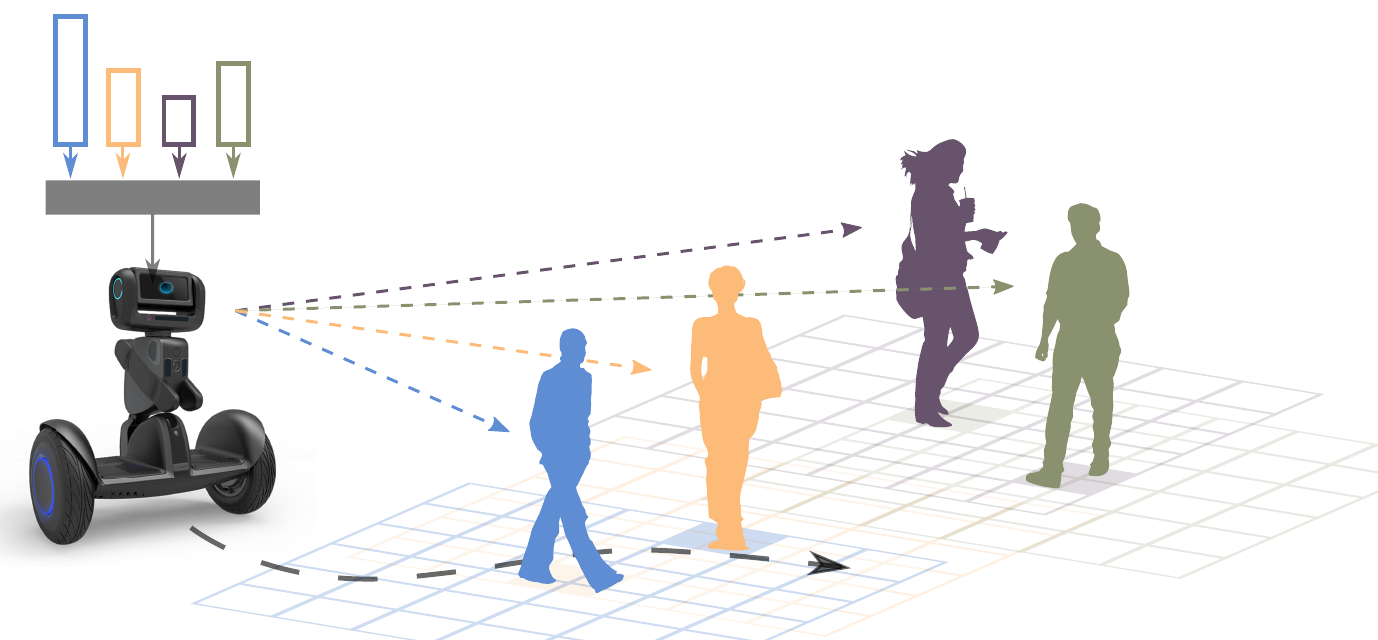}
  \caption{In this work, we present a method that jointly model Human-Robot and Human-Human interactions for navigation in crowds.}
  \label{fig:overview}
\end{figure}

In this work, we address the above issues by going beyond first-order Human-Robot interaction and dive into Crowd-Robot Interaction (CRI). We propose to: (i) rethink Human-Robot pairwise interactions with a self-attention mechanism, and (ii) jointly model Human-Robot as well as Human-Human interactions in the reinforcement learning framework. Inspired by \cite{alahi_social_2016,gupta_social_2018,vemula_social_2017}, our model extracts features for pairwise interactions between the robot and each human and captures the interactions among humans via local maps. Subsequently, we aggregate the interaction features with a self-attention mechanism that infers the relative importance of neighboring humans with respect to their future states. Our proposed model can naturally take into account an arbitrary number of agents, providing a good understanding of the crowd behavior for planning. An extensive set of simulation experiments shows that our approach can anticipate crowd dynamics and navigate in time efficient paths, outperforming state-of-the-art methods. We also demonstrate the effectiveness of our model on a robotic platform in real-world environments. The code of our approach is available at \texttt{\small https://github.com/vita-epfl/CrowdNav}. 

\section{BACKGROUND} \label{sec:background} 

\subsection{Related Work}
Earlier works have largely leveraged well-engineered interaction models to enhance the social awareness in robot navigation. One pioneering work is the Social Force \cite{helbing_social_1995-1,robicquet2016learning,kretz_indications_2018}, which has been successfully applied to autonomous robots in simulation and real-world environments \cite{sud_real-time_2007,ferrer_robot_2013,ferrer_robot_2017}. Another method named Interacting Gaussian Process (IGP) models the trajectory of each agent as an individual Gaussian Process and proposes an interaction potential term to couple the individual GP for interaction \cite{trautman_unfreezing_2010, trautman_robot_2013,trautman_sparse_2017}. In multi-agent settings, where the same policy is applied to all the agents, reactive methods such as RVO \cite{berg_reciprocal_2008} and ORCA \cite{van_den_berg_reciprocal_2011} seek joint obstacle avoidance velocities under reciprocal assumptions. The key challenge for these models is that they heavily rely on hand-crafted functions and cannot generalize well to various scenarios for crowd-like cooperation.

Another line of work uses imitation learning approaches to learn policies from demonstrations of desired behaviors. Navigation policies that map various inputs such as depth images, lidar measurements and local maps to control actions are developed in \cite{tai_socially_2017,long_deep-learned_2017,liu2018map} by directly mimicking expert demonstrations. Beyond behavioral cloning, inverse reinforcement learning has been used in \cite{roy_feature-based_2013, kretzschmar_socially_2016, pfeiffer_predicting_2016} to learn the underlying cooperation features from human data using the maximum entropy method. The learning outcomes in these works highly depend on the scale and quality of demonstrations, which is not only resource consuming but also constrains the quality of the learned policy by human efforts. In our work, we adopt the imitation learning approach to warm start our model training. 

Reinforcement Learning (RL) methods have been intensively studied over the last few years and applied to various fields since it started to achieve superior performance in video games \cite{mnih_human-level_2015}. In the field of robot navigation, recent works 
have used RL to learn sensorimotor policies in static and dynamic environments from the raw observations
\cite{tai_virtual--real_2017,long_towards_2017} and socially cooperative policies with the agent-level state information \cite{chen_decentralized_2016,chen_socially_2017,everett_motion_2018}. To handle a variant number of neighbors, the method reported in \cite{chen_decentralized_2016} adapts from the two-agent to the multi-agent case through a maximin operation that picks up the best action against the worst-case for the crowd. A later extension uses an LSTM model to process the state of each neighbor sequentially in reverse order of the distance to the robot \cite{everett_motion_2018}. In contrast to these simplifications, we propose a novel neural network model to capture the collective impact of the crowd explicitly. 

A variety of deep neural networks architectures have been proposed in recent years to improve the modeling of Human-Human interactions  \cite{alahi2017learning}. Handcrafted methods (\textit{e.g.}, based on Discrete Choice models \cite{antonini2006discrete}) have a nice interpretable property but have limited prediction power as recently shown in \cite{sifringer2018let}. Data-driven methods such as the Social LSTM method \cite{alahi_social_2016} models each individual by an LSTM and shares the states of neighboring LSTMs through a social pooling module. More recently, generative models are used for improved accuracy and efficiency \cite{fernando_soft_2017, gupta_social_2018, sadeghian_sophie:_2018}. Yet, the generative models still suffer from mode collapse \cite{liu2019collaborative}. Some other works model the social interactions through spatio-temporal graphs, where an attention model is introduced recently to learn the relative importance of each agent \cite{vemula_social_2017}. Sadeghian \textit{et al.} \cite{sadeghian2018car} study various attention mechanisms. In our work, that is built upon these models, we design a social attentive pooling module to encode crowd cooperative behaviors in a deep reinforcement learning framework. 


\subsection{Problem Formulation}
In this work, we consider a navigation task where a robot moves towards a goal through a crowd of $n$ humans. This can be formulated as a sequential
decision making problem in a reinforcement learning framework \cite{chen_decentralized_2016,chen_socially_2017,everett_motion_2018}. For each agent (robot or human), the position $\mathbf{p}=[p_x,p_y]$, velocity $\mathbf{v}=[v_x,v_y]$ and radius $r$ can be observed by the others. The robot is also aware of its unobservable state including the goal position $\mathbf{p_g}$ and preferred speed $v_{pref}$. We assume that the velocity of the robot $\mathbf{v}_t$ can be achieved instantly after the action command $\mathbf{a}_t$, \textit{i.e.}, $\mathbf{v}_t = \mathbf{a}_t$. Let $\mathbf{s}_t$ denote the state of the robot and $\mathbf{w}_t = [\mathbf{w}_t^1, \mathbf{w}_t^2, \dots, \mathbf{w}_t^n]$ denote the state of humans at time $t$. The joint state for robot navigation is defined as $\mathbf{s}_t^{jn} = [\mathbf{s}_t, \mathbf{w}_t]$. 

The optimal policy, $\pi^* : \mathbf{s}_t^{jn} \mapsto \mathbf{a}_t$, is to maximize the expected return:
\begin{equation} \label{eq:RL}
\begin{aligned}
\pi^{*}(\mathbf{s}_t^{jn}) = & \underset{\mathbf{a}_t}{\text{argmax}} ~ R(\mathbf{s}_t^{jn},\mathbf{a}_t) + \\
& \gamma^{\Delta t \cdot v_{pref}} \int_{\mathbf{s}_{t+\Delta t}^{jn}} P(\mathbf{s}_t^{jn},\mathbf{a_t},\mathbf{s}_{t+\Delta t}^{jn}) V^*(\mathbf{s}_{t+\Delta t}^{jn}) d\mathbf{s}_{t+\Delta t}^{jn} \\
V^*(\mathbf{s}_t^{jn}) = & \sum_{t'=t}^T \gamma^{t' \cdot v_{pref}} R_{t'}(\mathbf{s}_{t'}^{jn},\pi^*(\mathbf{s}_{t'}^{jn})), \\
\end{aligned}
\end{equation}
where $R_t(\mathbf{s}_t^{jn},\mathbf{a}_t)$ is the reward received at time $t$, $\gamma \in (0,1)$ is a discount factor, $V^*$ is the optimal value function, $P(\mathbf{s}_t^{jn},\mathbf{a_t},\mathbf{s}_{t+\Delta t}^{jn}) $ is the transition probability from time $t$ to time $t+\Delta t$. The preferred velocity $v_{pref}$ is used as a normalization term in the discount factor for numerical reasons \cite{chen_decentralized_2016}.

We follow the formulation of the reward function defined in \cite{chen_decentralized_2016,chen_socially_2017}, which awards task accomplishments while penalizing collisions or uncomfortable distances, 
\begin{equation}
    R_t(\mathbf{s}_t^{jn},\mathbf{a}_t)= 
\begin{dcases}
    -0.25 & \text{if} ~~ d_t < 0 \\
    -0.1 + d_t/2 & \text{else if} ~~ d_t < 0.2 \\
    1 & \text{else if} ~~ \mathbf{p}_t=\mathbf{p}_g \\
    0 & \text{otherwise} 
\end{dcases}
\end{equation}
where $d_t$ is the minimum separation distance between the robot and the humans during the time period $[t-\Delta t,t]$. 

\subsection{Value Network Training}

The value network is trained by the temporal-difference method with standard experience replay and fixed target network techniques \cite{mnih_human-level_2015,chen_decentralized_2016}. As outlined in Algorithm \ref{alg:v-learning}, the model is first initialized with imitation learning using a set of demonstrator experiences (line \ref{line:init-v}-\ref{line:init-memory}), and subsequently refined from experience of interactions (line \ref{line:rl-start}-\ref{line:rl-end}). One distinction from the previous works \cite{chen_decentralized_2016,chen_socially_2017} is that the next state $S_{t+1}^{jn}$ in line \ref{line:action_selection} is obtained by querying the environment the true value instead of approximating with a linear motion model, mitigating the issue of system dynamics in training. During deployment, the transition probability can be approximated by a trajectory prediction model \cite{helbing_social_1995,alahi_social_2016,gupta_social_2018}. 

\begin{algorithm} [t]
\small
\captionsetup{font=small}
\caption{Deep V-learning}
\begin{algorithmic}[1] 
\State Initialize value network $V$ with demonstration $\mathsf{D}$ \label{line:init-v}
\State Initialize target value network $\hat{V} \leftarrow V$ 
\State Initialize experience replay memory $\mathsf{E} \leftarrow \mathsf{D}$ \label{line:init-memory}
\For{episode = 1, M} \label{line:rl-start}
\State Initialize random sequence $\mathbf{s}_0^{jn}$ 
\Repeat
\State $a_t \leftarrow argmax_{a_t \in A} R(s_t^{jn}, a_t) + \gamma^{\Delta t \cdot v_{pref}} V(_{t+\Delta t}^{jn})$ \label{line:action_selection}
\State Store tuple ($\mathbf{s}_t^{jn}, \mathbf{a}_t, r_t, \mathbf{s}_{t+\Delta t}^{jn}$) in $\mathsf{E}$ 
\State Sample random minibatch tuples from $\mathsf{D}$
\State Set target $y_i = r_i + \gamma^{\Delta t \cdot v_{pref}}  \hat{V}(\mathbf{s}_{i+1}^{jn})$
\State Update value network $V$ by gradient descent
\Until terminal state $\mathbf{s}_t$ or $ t \ge t_{max}$
\State Update target network $\hat{V} \leftarrow V$ 
\EndFor \label{line:rl-end}
\State \textbf{return} $V$
\end{algorithmic}
\label{alg:v-learning}
\end{algorithm}

To tackle the problem (\ref{eq:RL}) effectively, the value network model needs to accurately approximate the optimal value function $V^{*}$ that implicitly encodes the social cooperation among agents. Previous works on this track didn't fully model the crowd interactions, which degrades the accuracy of value estimation for a densely populated scene. In the following sections, we will present a novel Crowd-Robot Interaction model that can effectively learn to navigate in crowded spaces. 

\section{APPROACH} \label{sec:approach} 

When humans walk in a densely populated scene, they cooperate with others by anticipating the behaviors of their neighbors in the vicinity, particularly those who are likely to be involved in some future interactions. This motivates us to design a model that can calculate the relative importance and encode the collective impact of neighboring agents for socially compliant navigation. Inspired by the social pooling \cite{alahi_social_2016,gupta_social_2018} and attention models \cite{liu_learning_2016,vaswani_attention_2017,lin_structured_2017,vemula_social_2017,hoshen_vain:_2017,zhang_self-attention_2018}, we introduce a socially attentive network that consists of three modules: 

\begin{itemize}
    \item \textbf{Interaction module}: models the Human-Robot interactions explicitly and encodes the Human-Human interactions through coarse-grained local maps.
    \item \textbf{Pooling module}: aggregates the interactions into a fixed-length embedding vector by a self-attention mechanism.
    \item \textbf{Planning module}: estimates the value of the joint state of the robot and crowd for social navigation.
\end{itemize}

In the following subsections, we present the architecture and formulations of each module. The time index $t$ is omitted below for simplicity. 

\begin{figure}[tb]
  \centering
  \captionsetup{font=small}
  \includegraphics[width=0.5\textwidth]{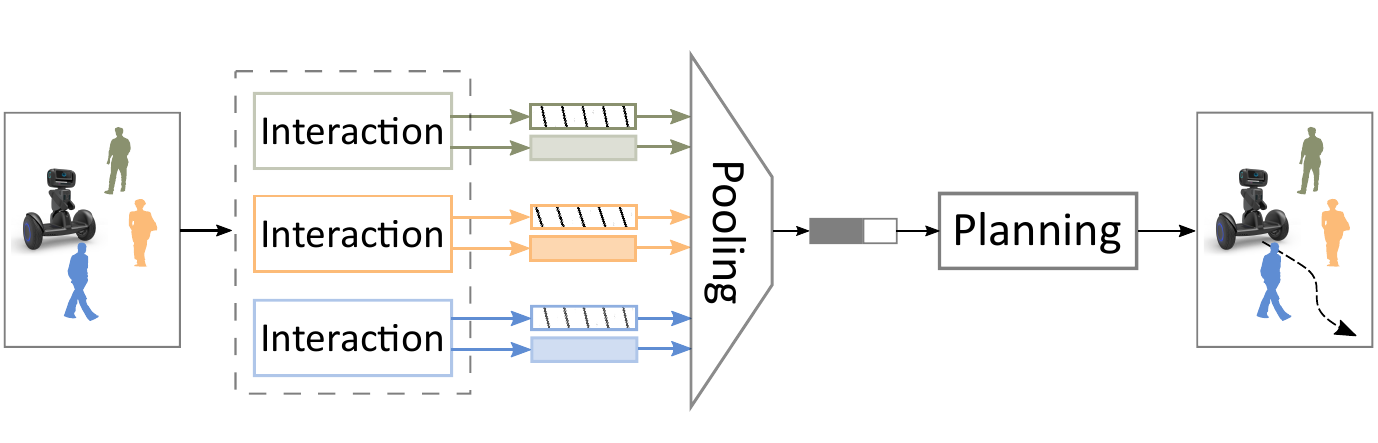}
  \caption{Overview of our method for socially attentive navigation made of 3 modules: Interaction, Pooling, and Planning described in Section \ref{sec:approach}. Interactions between the robot and each human are extracted from the interaction module and subsequently aggregated in the pooling module. The planning module estimates the value of the joint state of the robot and humans for navigation in crowds.}
  \label{fig:model}
\end{figure}

\subsection{Parameterization}


We follow the robot-centric parameterization in \cite{chen_decentralized_2016, everett_motion_2018}, where the robot is located at the origin and the x-axis is pointing toward the robot's goal. The states of the robot and walking humans after transformation are: 
\begin{equation}
\begin{aligned} \label{eq:parameterization}
    s &= [d_g, v_{pref}, v_x, v_y, r], \\
    {w}_i &= [{p}_x, {p}_y, {v}_x, {v}_y, {r}_i, {d}_i, {r}_i + r],
\end{aligned}{}
\end{equation}
where $d_g=||\mathbf{p} - \mathbf{p}_g||_2$ is the robot's distance to the goal and $d_i=||\mathbf{p} - {\mathbf{p}_i}||_2$ is the robot's distance to the neighbor $i$.

\subsection{Interaction Module}
Each human has an impact on the robot and is meanwhile influenced by his/her neighboring humans. Explicitly modeling all pairs of interactions among humans leads to $O(N^2)$ complexity \cite{vemula_social_2017}, which is not computationally-desirable for a policy to scale up in dense scenes. We tackle this problem by introducing a pairwise interaction module that explicitly models the Human-Robot interaction while using local maps as a coarse-grained representation for the Human-Human interactions.

\begin{figure}[tb]
  \centering
  \captionsetup{font=small}
  \includegraphics[width=0.48\textwidth]{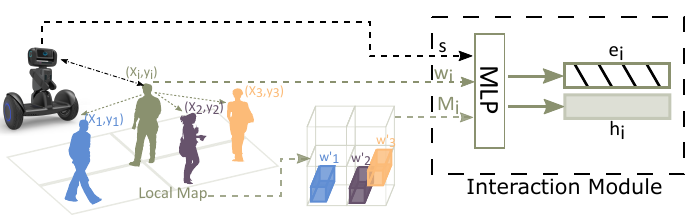}
  \caption{Illustration of our interaction module. We use a multi-layer perceptron to extract the pairwise interaction feature between the robot and each human $i$. The impact of the other people on the human $i$ is represented by a local map.}
  \label{fig:pairwise}
\end{figure}

Given a neighborhood of size $L$, we construct a $L \times L \times 3$ map tensor $M_i$ centered at each human $i$ to encode the presence and velocities of neighbors, which is referred as local map in Fig.~\ref{fig:pairwise}: 
\begin{equation} \label{eq:om}
    M_i(a,b,:) = \sum_{j \in \mathcal{N}_i} \delta_{ab}[x_j-x_i,y_j-y_i]w^{\prime}_j,
\end{equation}
where $w^{\prime}_j = (v_{xj},v_{yj},1)$ is a local state vector for human $j$, $\delta_{mn}[x_j-x_i,y_j-y_i]$ is an indicator function which equals to $1$ only if the relative position $(\Delta x, \Delta y)$ is located in the cell $(a,b)$, $ \mathcal{N}_i$ is the set of neighboring humans around the $i^{th}$ person. 

We embed the state of human $i$ and the map tensor $M_i$, together with the state of the robot, into a fixed length vector $e_i$ using a multi-layer perceptron (MLP): 
\begin{equation} \label{eq:embed}
    e_i = \phi_e(s, w_i, M_i; W_e),
\end{equation}
where $\phi_e(\cdotp)$ is an embedding function with ReLU activations and $W_e$ is the embedding weights.

The embedding vector $e_i$ is fed to a subsequent MLP to obtain the pairwise interaction feature between the robot and person $i$:
\begin{equation} \label{eq:interaction}
    h_i = \psi_h(e_i; W_h),
\end{equation}
where $\psi_h(\cdotp)$ is a fully-connected layer with ReLU non-linearity and $W_h$ is the network weights. 

\subsection{Pooling Module}
Since the number of surrounding humans can vary dramatically in different scenes, we need a model that can handle an arbitrary number of inputs into a fixed size output. Everett et al. \cite{everett_motion_2018} proposed to feed the states of all humans into an LSTM \cite{hochreiter_long_1997} sequentially in descending order of their distances to the robot. However, the underlying assumption that the closest neighbors have the strongest influence is not always true. Some other factors, such as speed and direction, are also essential for correctly estimating the importance of a neighbor, which reflects how this neighbor could potentially influence the robot's goal acquisition. Leveraging the recent progress in the self-attention mechanism, where the attention of an item in a sequence is gained by looking at other items in the sequence \cite{liu_learning_2016,lin_structured_2017,conneau_supervised_2017}, we propose a social attentive pooling module to learn the relative importance of each neighbor and the collective impact of the crowd in a data-driven fashion. 

The interaction embedding $e_i$ is transformed into an attention score $\alpha_i$ as follows: 
\begin{equation} \label{eq:mean-pooling}
    e_m = \frac{1}{n} \sum_{k=1}^{n} e_k,
\end{equation}
\begin{equation} \label{eq:score}
    \alpha_i = \psi_\alpha(e_i, e_m; W_\alpha),
\end{equation}
where $e_m$ is a fixed-length embedding vector obtained by mean pooling all the individuals,  $\psi_\alpha(\cdotp)$ is an MLP with ReLU activations and $W_\alpha$ are the weights. 

Given the pairwise interaction vector $h_i$ and the corresponding attention score $\alpha_i$ for each neighbor $i$, the final representation of the crowd is a weighted linear combination of all the pairs: 
\begin{equation}
    c = \sum_{i=1}^{n} \textrm{softmax}(\alpha_i) h_i.
\end{equation}

\begin{figure}[tb]
  \centering
  \captionsetup{font=small}
  \includegraphics[width=0.48\textwidth]{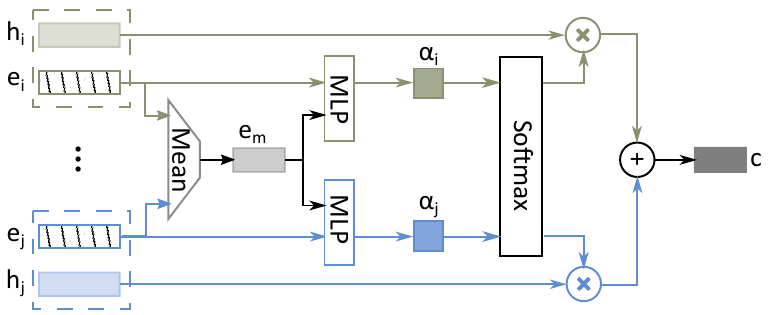}
  \caption{Architecture of our pooling module. We use a multi-layer perceptron to compute the attention score for each person from the individual embedding vector together with the mean embedding vector. The final joint representation is a weighted sum of the pairwise interactions.}
  \label{fig:pooling}
\end{figure}

\subsection{Planning Module}

Based on the compact representation of the crowd $c$, we build a planning module that estimates the state value $v$ for cooperative planning: 


\begin{equation} \label{eq:planning}
v = f_v(s,c; W_v),
\end{equation}
where $f_v(\cdotp)$ is an MLP with ReLU activations, the weights are denoted by $W_v$. 

\subsection{Implementation Details} 
The local map is a $4 \times 4$ grid centered at each human and the side length of each cell is $1m$ . The hidden units of functions $\phi_e(\cdotp), \psi_h(\cdotp), \psi_\alpha(\cdotp), f_v(\cdotp)$ are  (150,100), (100,50), (100,100), (150,100,100) respectively. 

We implemented the policy in PyTorch \cite{paszke_automatic_2017} and trained it with a batch size of $100$ using Adam \cite{kingma_adam:_2014}. For imitation learning, we collected $3k$ episodes demonstration using ORCA and trained the policy 50 epochs with learning rate $0.01$. For reinforcement learning, the learning rate is $0.001$ and the discount factor $\gamma$ is 0.9.
The exploration rate of the $\epsilon$-greedy policy decays linearly from 0.5 to 0.1 in the first $5k$ episodes and stays 0.1 for the remaining $5k$ episodes. The RL training took approximately 10 hours on an i7-8700 CPU.

This work assumes holonomic kinematics for the robot, \textit{i.e.}, it can move in any direction. The action space consists of 80 discrete actions: 5 speeds exponentially spaced between (0, $v_{pref}$] and 16 headings evenly spaced between [0, 2$\pi$).

\section{EXPERIMENTS}
\subsection{Simulation Setup}
We built a simulation environment in Python for robot navigation in crowds. The simulated humans are controlled by ORCA \cite{van_den_berg_reciprocal_2011}, the parameters of which are sampled from a Gaussian distribution to introduce behavioral diversity. We use circle crossing scenarios for both training and test, where all the humans are randomly positioned on a circle of radius $4m$ and their goal positions are on the opposite side of the same circle. Random perturbation is added to x,y coordinates of both starting and goal positions.

Three existing state-of-the-art methods, ORCA \cite{berg_reciprocal_2008}, CADRL \cite{chen_decentralized_2016} and LSTM-RL \cite{everett_motion_2018}, are implemented as baseline methods. The main distinction between our method and the RL baselines lies in the interaction and pooling module, we keep the planning module identical for a fair comparison. Note that the LSTM-RL in our implementation differs from the original one \cite{everett_motion_2018} in that we use the joint state instead of human's observable state as the input of the LSTM unit. We refer to our full model as LM-SARL and the model without a local map as SARL for ablation experiments.  

To fully evaluate the effectiveness of the proposed model, we look into two simulation settings: invisible and visible. The former one sets the robot invisible to the other humans. As a result, the simulated humans react only to humans but not to the robot. We also removed the penalty on the uncomfortable distance in the reward function to eliminate extra factors for collision avoidance. This setting serves as a clean testbed for validating the model's ability in reasoning the Human-Robot and Human-Human interaction without affecting human's behaviors. The latter visible setting resembles more realistic cases where the robot and humans have mutual impacts. Models are evaluated with 500 random test cases in both settings.

\subsection{Quantitative Evaluation} \label{sec:quanti}
\subsubsection{Invisible Robot} 
In the invisible setting, a robot needs to forecast future trajectories of all the humans to avoid collisions. Table~\ref{tab:invisible} reports the rates of success, collision, the average navigation time as well as the average discounted cumulative reward in test experiments. 

As expected, the ORCA method fails badly in the invisible setting due to the violation of the reciprocal assumption. Among all the reinforcement learning methods, the CADRL has the lowest success rate. This is because the maximin approach used in the CADRL can only take a single pair of interaction into account while ignoring the rest. The frequent failure of the CADRL shows the necessity for a policy to take all humans into account simultaneously.

By directly aggregating the surrounding agents' information, both LSTM-RL and SARL achieve a higher success rate. However, LSTM-RL suffers from occasional collisions and timeouts, whereas the SARL accomplishes all the test cases. We also observe a dramatic reduction in the average navigation time in the SARL. These results demonstrate the advantages of the proposed attentive pooling mechanism in capturing the collective impact of the crowd. The full version of our model, LM-SARL, achieves the best results in the invisible experiments, outperforming the SARL in terms of both the navigation time and the cumulative reward. Though not by a large margin, this improvement indicates the benefits of encoding the interactions among humans. 

\subsubsection{Visible Robot} We further compare the navigation performance of our models with the baselines in the visible setting. The robot not only needs to understand the behavior of humans but also interact with them to obtain high rewards. We define the discomfort frequency as $t_\text{disc} / {T}$, where $t_\text{disc}$ is the duration when the separation distance $d_t<$0.2m. To compare the ORCA baseline with the learning methods fairly, we add an extra 0.1m as the virtual radius of the agent to maintain a comfortable distance to humans for human-aware navigation \cite{kruse_human-aware_2013}. The results are summarized in Table~\ref{tab:visible}. 

Different from the invisible case which violates the reciprocal assumption, the ORCA policy in the visible setting achieves a high success rate and never invades the comfort zone of the other humans. However, the ORCA baseline fails to obtain high rewards due to the short-sighted and conservative behaviors. As pointed out in \cite{long_deep-learned_2017}, tuning ORCA towards an objective function can be a tedious and challenging process compared with learning-based methods.

The Reinforcement Learning results in the visible setting are similar to the invisible ones as expected. Our SARL model outperforms the baselines significantly, and the LM-SARL shows further improvements on the final reward. Since the Human-Human interactions are not significant all the time, their effect on the quantitative results is diluted over episodes. However, we see qualitative improvements which we discuss in the next section. 

\begin{table}[t]
\centering
\begin{tabular}{|c|c|c|c|c|}
\hline
Methods     & Success       & Collision         & Time & Reward          \\ \hline
ORCA \cite{berg_reciprocal_2008}        & 0.43          & 0.57          & 10.86         & 0.054         \\ \hline
CADRL \cite{chen_decentralized_2016}      & 0.78          & 0.22          & 10.80         & 0.222         \\ \hline
LSTM-RL \cite{everett_motion_2018}    & 0.95          & 0.03          & 11.82         & 0.279         \\ \hline
SARL (Ours)     & \textbf{1.00} & \textbf{0.00} & 10.55         & 0.338         \\ \hline
LM-SARL (Ours)    & \textbf{1.00} & \textbf{0.00} & \textbf{10.46}& \textbf{0.342}\\
\hline
\end{tabular}
\captionsetup{font=small}
\caption{Quantitative results in the invisible setting. ``Success":~the rate of robot reaching its goal without a collision. ``Collision":~the rate of robot colliding with other humans. ``Time":~the robot's navigation time to reach its goal in seconds. ``Reward":~discounted cumulative reward in a navigation task.}
\label{tab:invisible}
\end{table}

\begin{table}[t]
\begin{tabularx}{\linewidth}{|c|c|c|c|c|X|}
\hline
Methods     & Success       & Collision  & Time & Disc.  & Reward          \\ \hline
ORCA \cite{berg_reciprocal_2008} & 0.99 & 0.00* & 12.29 & 0.00* & 0.284 \\ \hline
CADRL \cite{chen_decentralized_2016} & 0.94 & 0.03 & 10.82 & 0.10 & 0.291 \\ \hline
LSTM-RL \cite{everett_motion_2018} & 0.98 & 0.02 & 11.29 & 0.05 & 0.299 \\ \hline
SARL (Ours) & 0.99 & 0.01 & \textbf{10.58} & \textbf{0.02}  & 0.332 \\ \hline
LM-SARL (Ours) & \textbf{1.00} & \textbf{0.00} & 10.59 & 0.03 &  \textbf{0.334} \\
\hline
\end{tabularx}
\captionsetup{font=small}
\caption{Quantitative results in the visible setting. ``Disc." refers to as the discomfort  frequency (\% of duration where robot is too close to other humans). (*) Note that ORCA has a ``Collision" and ``Disc." of 0 by design.}
\label{tab:visible}
\end{table}

\subsection{Qualitative Evaluation}

\begin{figure}[tb]
\captionsetup{font=small}
\captionsetup[sub]{font=small}
\begin{subfigure}{0.24\textwidth}
\includegraphics[width=\linewidth]{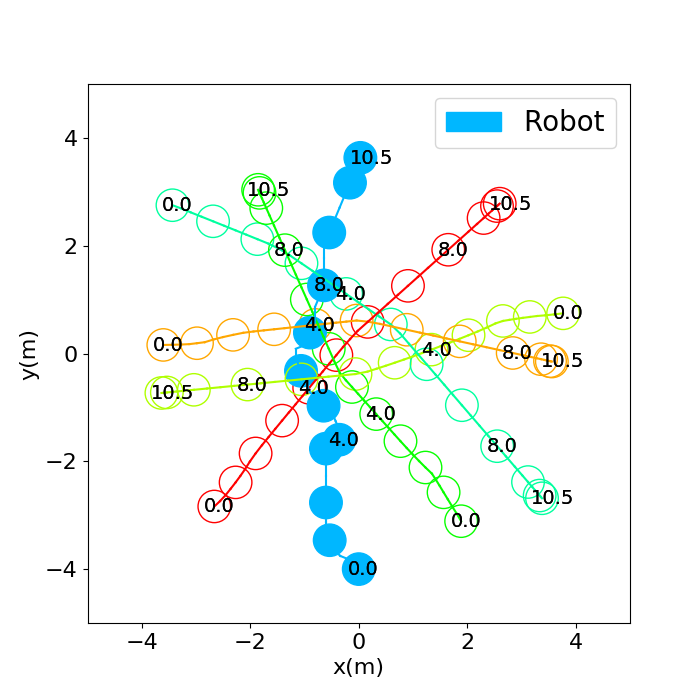}
\caption{CADRL \cite{chen_decentralized_2016}}
\label{fig:cadrl_traj}
\end{subfigure}
\hfill
\begin{subfigure}{0.24\textwidth}
\includegraphics[width=\linewidth]{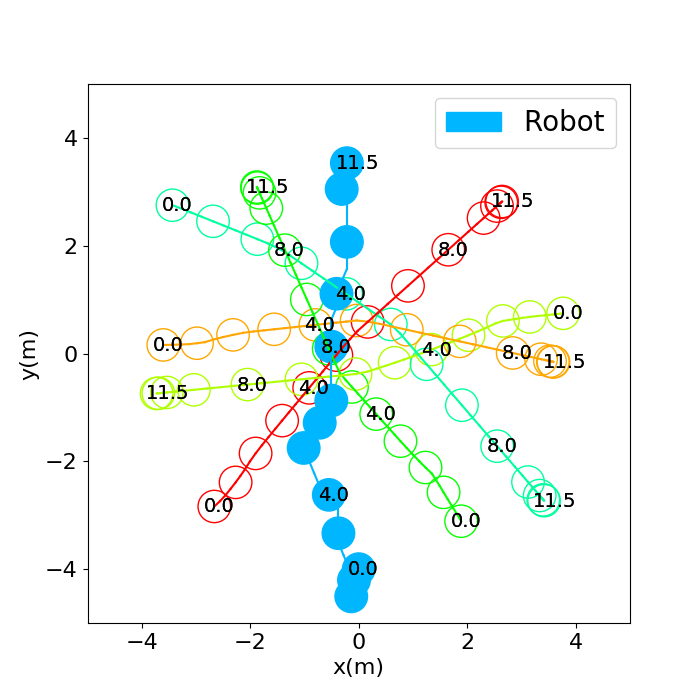}
\caption{LSTM-RL \cite{everett_motion_2018}}
\label{fig:lstm_rl_traj}
\end{subfigure}
\hfill
\begin{subfigure}{0.24\textwidth}
\includegraphics[width=\linewidth]{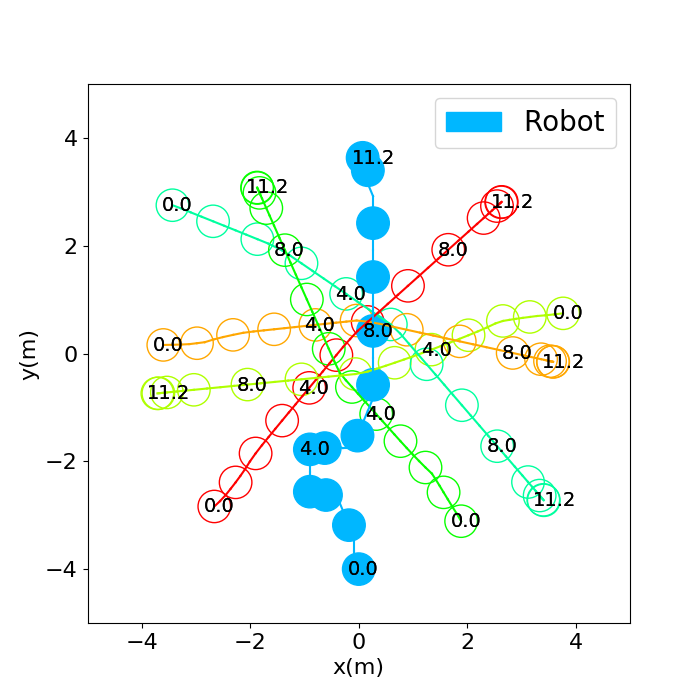}
\caption{Our SARL}
\label{fig:sarl_traj}
\end{subfigure}
\hfill
\begin{subfigure}{0.24\textwidth}
\includegraphics[width=\linewidth]{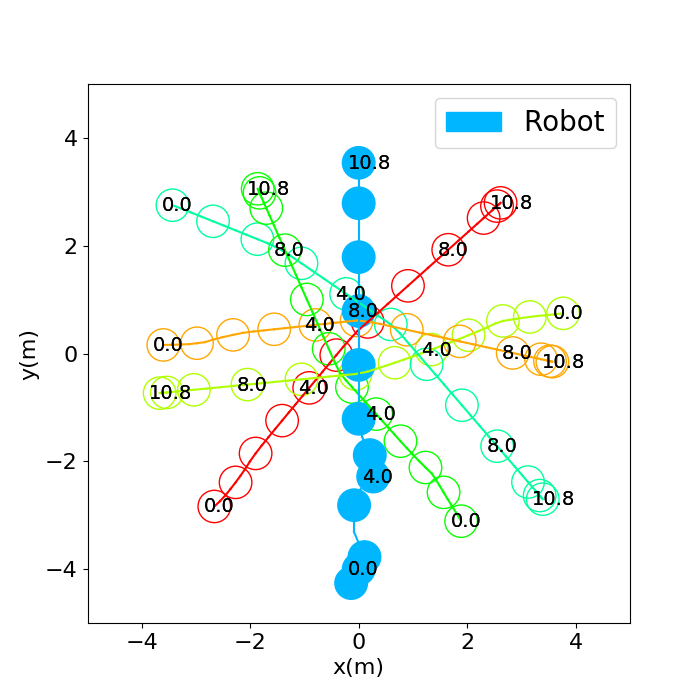}
\caption{Our LM-SARL}
\label{fig:om_sarl_traj}
\end{subfigure}
\caption{Trajectory comparison in an invisible test case. Circles are the positions of agents at the labeled times. When encountering humans, CADRL and LSTM-RL demonstrate overly aggressive and conservative behaviors respectively. In contrast, our SARL and LM-SARL successfully identify a shortcut through the center, which allows the robot to keep some distance from others while navigating to the goal quickly.}
\label{fig:trajectory}
\end{figure}

We further investigate the effectiveness of our model through qualitative analysis. As shown in Fig. \ref{fig:trajectory}, the navigation paths of different methods are compared in an invisible test case, where the trajectories of humans are identical for a clear comparison. When encountering humans in the center of the space, the CADRL passes them aggressively. By contrast, the LSTM-RL slows down dramatically to avoid the crowd from 4.0s to 8.0s, ending up with a long navigation time. In comparison to the baselines, our SARL hesitates at first but then recognizes a shorter path to the goal through the center. By taking the shortcut, the robot successfully avoids other humans. The LM-SARL identifies the central highway from the very beginning, establishing a smart trace concerning both the safety distance and navigation time. 

\begin{figure} [tb]
  \centering
  \captionsetup{font=small}
  \includegraphics[width=0.24\textwidth]{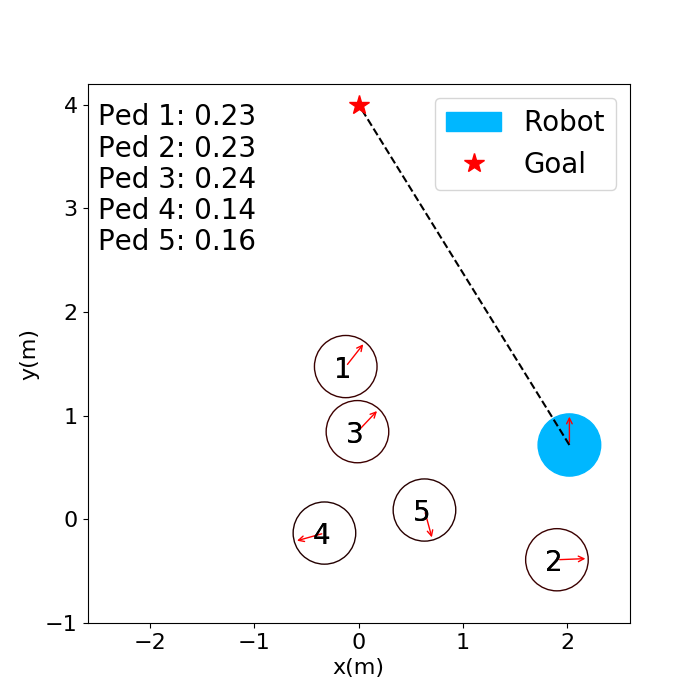}
  \caption{Attention scores in a dense scene. Our LM-SARL assigns low importance scores to human $\#$4 and $\#$5 who walk away, whereas attending with the highest weight to $\#$3 who is most likely to get close soon.}
  \label{fig:attention}
\end{figure}

In addition to the overall trajectory, we take a closer look at the learned policies in a typical crowded frame. Fig.~\ref{fig:attention} shows the attention scores of humans inferred by our LM-SARL model. The lowest attention score is assigned to the~$\#$4 human who has the largest distance to the robot. The human~$\#$5 located not far from the robot also receives a low score, as he is walking away from the robot. In contrast, our model pays more attention to $\#$1, $\#$2 and $\#$3, all of which have a potential influence on the robot's path planning. Human~$\#$2 is the closest to the robot and obtains high attention score. However, our model gives the highest attention score to $\#$3, who is most likely to get closest to the robot in the next few steps. Through assigning importance scores to humans, our attentive pooling module demonstrates a good ability to reason the relative importance of humans in a dense scene. 

\begin{figure}[t]
\captionsetup{font=small}
\captionsetup[sub]{font=small}
\begin{subfigure}{0.24\textwidth}
\captionsetup[sub]{font=small}
\includegraphics[width=\linewidth]{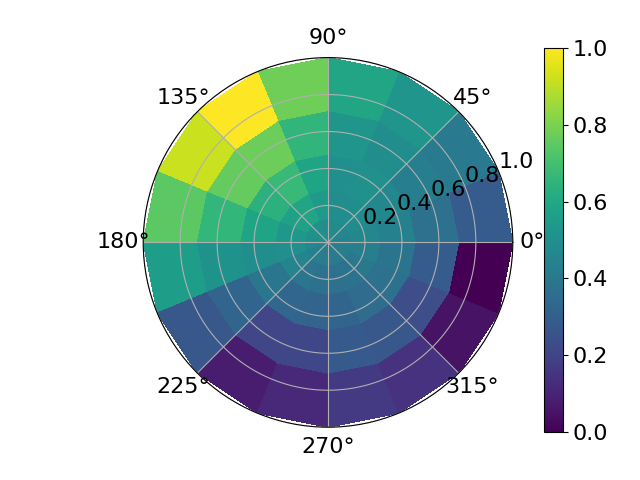}
\caption{CADRL \cite{chen_decentralized_2016}}
\end{subfigure}
\hfill
\begin{subfigure}{0.24\textwidth}
\includegraphics[width=\linewidth]{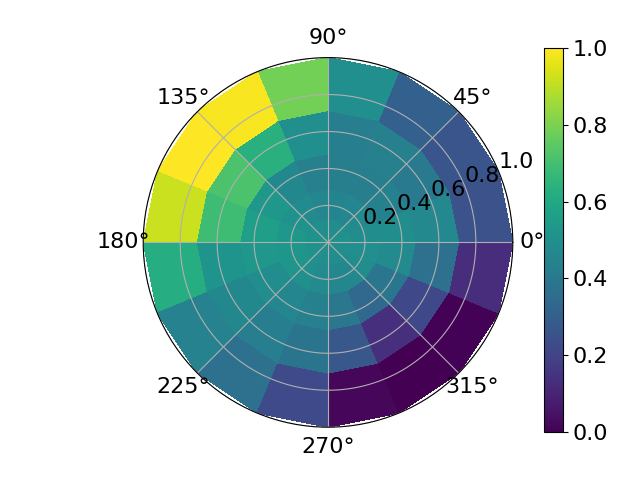}
\caption{LSTM-RL \cite{everett_motion_2018}}
\end{subfigure}
\hfill
\begin{subfigure}{0.24\textwidth}
\includegraphics[width=\linewidth]{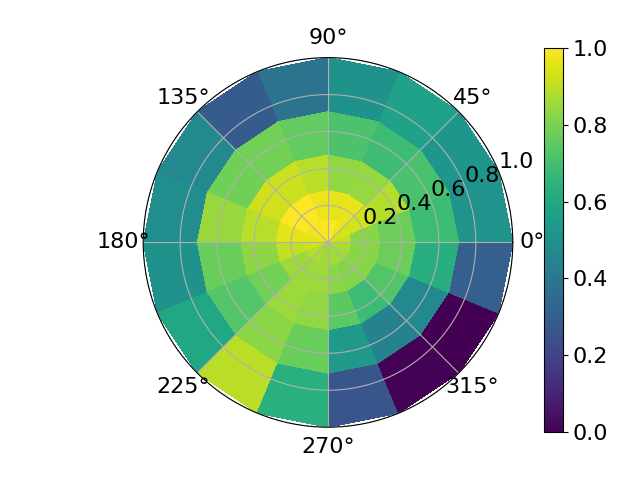}
\caption{Our SARL}
\end{subfigure}
\hfill
\begin{subfigure}{0.24\textwidth}
\includegraphics[width=\linewidth]{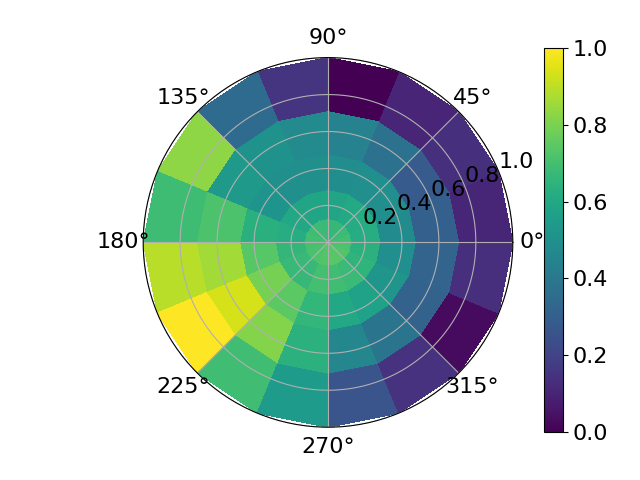}
\caption{Our LM-SARL}
\end{subfigure}
\caption{Value estimations by different methods for the dense scene in Fig.~\ref{fig:attention}. The baseline methods predict high values for high speeds towards the goal, which is dangerous because of humans $\#$1 and $\#$3. In contrast, our SARL slows down and waits safely, and our LM-SARL prefers to turn to 200$^\circ$, preparing to pass behind them.}
\label{fig:vf}
\end{figure}

As the ultimate objective of our model is to accurately estimate the state value, we finally compare the values estimated by different methods in Fig.~\ref{fig:vf}. Given that humans $\#1$ and $\#3$ are likely to cross the straight path from the robot to the goal, the robot is expected to either step aside or to slow down to avoid them. 

Limited by the maximin selection, CADRL predicts low values only in the direction towards the closest human $\#$2 while erroneously assigning the highest value to the direction $120^{\circ}$. The LSTM-RL model shifts the action preference slightly to the left but still overestimates the values of the high speeds in the directions around $120^{\circ}$.

In contrast, our SARL model predicts distinctly low values for the full speeds in the dangerous directions from $70^{\circ}$ to $200^{\circ}$, leading to a slow down to avoid collisions. Considering the social repulsive forces between the person $\#1$ and $\#3$, person $\#3$ might turn to the robot in the future, raising potential dangers or delays in the $120^{\circ}$ direction. By encoding the Human-Human interactions through local maps, LM-SARL succeeds in providing a smart action in the $200^{\circ}$ direction, which paves the way for cutting behind $\#1$ and $\#3$. This indicates LM-SARL's potential for reasoning about complex interactions among agents. 

\subsection{Real-world Experiments}
Aside from the simulation experiments above, we also examine the trained policy in real-world experiments on a Segway robotic platform and the video demo can be found at \texttt{\small https://youtu.be/0sNVtQ9eqjA}. 



\section{CONCLUSION}
In this work, we tackle the crowd navigation problem by decomposing the Crowd-Robot Interaction into two parts. We first jointly model the Human-Robot and Human-Human interactions and then aggregate the interactions into a compact crowd representation via a self-attention model. Our approach outperforms state-of-the-art navigation methods in terms of time-efficiency and task accomplishments. Qualitatively, we demonstrate our model's ability to reason the importance of humans in a crowd.

\textbf{Acknowledgements}. We acknowledge the support of Samsung and Segway Robotics for the Loomo hardware. We also thank Farshid Moussavi from Samsung for helpful discussions.



\addtolength{\textheight}{-0cm}   



\bibliographystyle{IEEEtran}
\bibliography{vita-social-nav}

\begin{thebibliography}{10}
\providecommand{\url}[1]{#1}
\csname url@samestyle\endcsname
\providecommand{\newblock}{\relax}
\providecommand{\bibinfo}[2]{#2}
\providecommand{\BIBentrySTDinterwordspacing}{\spaceskip=0pt\relax}
\providecommand{\BIBentryALTinterwordstretchfactor}{4}
\providecommand{\BIBentryALTinterwordspacing}{\spaceskip=\fontdimen2\font plus
\BIBentryALTinterwordstretchfactor\fontdimen3\font minus
  \fontdimen4\font\relax}
\providecommand{\BIBforeignlanguage}[2]{{%
\expandafter\ifx\csname l@#1\endcsname\relax
\typeout{** WARNING: IEEEtran.bst: No hyphenation pattern has been}%
\typeout{** loaded for the language `#1'. Using the pattern for}%
\typeout{** the default language instead.}%
\else
\language=\csname l@#1\endcsname
\fi
#2}}
\providecommand{\BIBdecl}{\relax}
\BIBdecl
\renewcommand{\BIBentryALTinterwordstretchfactor}{4}

\bibitem{borenstein_real-time_1989}
J.~Borenstein and Y.~Koren, ``Real-time obstacle avoidance for fast mobile
  robots,'' \emph{IEEE Transactions on Systems, Man, and Cybernetics}, vol.~19,
  no.~5, pp. 1179--1187, Sep. 1989.

\bibitem{borenstein_real-time_1990}
J.~Borenstein and Y.~Koren, ``Real-time obstacle avoidance for fast mobile
  robots in cluttered environments,'' in \emph{, {IEEE} {International}
  {Conference} on {Robotics} and {Automation} {Proceedings}}, May 1990, pp.
  572--577 vol.1.

\bibitem{borenstein_vector_1991}
J.~Borenstein and Y.~Koren, ``The vector field histogram-fast obstacle
  avoidance for mobile robots,'' \emph{IEEE Transactions on Robotics and
  Automation}, vol.~7, no.~3, pp. 278--288, Jun. 1991.

\bibitem{fox_dynamic_1997}
D.~Fox, W.~Burgard, and S.~Thrun, ``The dynamic window approach to collision
  avoidance,'' \emph{IEEE Robotics Automation Magazine}, vol.~4, no.~1, pp.
  23--33, Mar. 1997.

\bibitem{berg_reciprocal_2008}
J.~v.~d. Berg, M.~Lin, and D.~Manocha, ``Reciprocal {Velocity} {Obstacles} for
  real-time multi-agent navigation,'' in \emph{2008 {IEEE} {International}
  {Conference} on {Robotics} and {Automation}}, May 2008, pp. 1928--1935.

\bibitem{van_den_berg_reciprocal_2011}
J.~van~den Berg, S.~J. Guy, M.~Lin, and D.~Manocha,
  ``\BIBforeignlanguage{en}{Reciprocal n-{Body} {Collision} {Avoidance}},'' in
  \emph{\BIBforeignlanguage{en}{Robotics {Research}}}, ser. Springer {Tracts}
  in {Advanced} {Robotics}, C.~Pradalier, R.~Siegwart, and G.~Hirzinger,
  Eds.\hskip 1em plus 0.5em minus 0.4em\relax Springer Berlin Heidelberg, 2011,
  pp. 3--19.

\bibitem{snape_hybrid_2011}
J.~Snape, J.~v.~d. Berg, S.~J. Guy, and D.~Manocha, ``The {Hybrid} {Reciprocal}
  {Velocity} {Obstacle},'' \emph{IEEE Transactions on Robotics}, vol.~27,
  no.~4, pp. 696--706, Aug. 2011.

\bibitem{fong_survey_2003}
T.~Fong, I.~Nourbakhsh, and K.~Dautenhahn, ``A survey of socially interactive
  robots,'' \emph{Robotics and Autonomous Systems}, vol.~42, no.~3, pp.
  143--166, Mar. 2003.

\bibitem{kruse_human-aware_2013}
T.~Kruse, A.~K. Pandey, R.~Alami, and A.~Kirsch, ``Human-aware robot
  navigation: {A} survey,'' \emph{Robotics and Autonomous Systems}, vol.~61,
  no.~12, pp. 1726--1743, Dec. 2013.

\bibitem{roy_feature-based_2013}
N.~Roy, P.~Newman, and S.~Srinivasa, ``Feature-{Based} {Prediction} of
  {Trajectories} for {Socially} {Compliant} {Navigation},'' in \emph{Robotics:
  {Science} and {Systems} {VIII}}.\hskip 1em plus 0.5em minus 0.4em\relax MITP,
  2013.

\bibitem{kretzschmar_socially_2016}
H.~Kretzschmar, M.~Spies, C.~Sprunk, and W.~Burgard,
  ``\BIBforeignlanguage{en}{Socially compliant mobile robot navigation via
  inverse reinforcement learning},'' \emph{\BIBforeignlanguage{en}{The
  International Journal of Robotics Research}}, vol.~35, no.~11, pp.
  1289--1307, Sep. 2016.

\bibitem{helbing_social_1995}
D.~Helbing and P.~Molnár, ``Social force model for pedestrian dynamics,''
  \emph{Physical Review E}, vol.~51, no.~5, pp. 4282--4286, May 1995.

\bibitem{alahi_social_2016}
A.~Alahi \emph{et~al.}, ``Social {LSTM}: {Human} {Trajectory} {Prediction} in
  {Crowded} {Spaces},'' in \emph{2016 {IEEE} {Conference} on {Computer}
  {Vision} and {Pattern} {Recognition} ({CVPR})}, Jun. 2016, pp. 961--971.

\bibitem{vemula_social_2017}
A.~Vemula, K.~Muelling, and J.~Oh, ``Social {Attention}: {Modeling} {Attention}
  in {Human} {Crowds},'' \emph{arXiv:1710.04689 [cs]}, Oct. 2017, arXiv:
  1710.04689.

\bibitem{gupta_social_2018}
A.~Gupta \emph{et~al.}, ``Social {GAN}: {Socially} {Acceptable} {Trajectories}
  with {Generative} {Adversarial} {Networks},'' \emph{arXiv:1803.10892 [cs]},
  Mar. 2018, arXiv: 1803.10892.

\bibitem{bennewitz_learning_2005}
M.~Bennewitz, W.~Burgard, G.~Cielniak, and S.~Thrun,
  ``\BIBforeignlanguage{en}{Learning {Motion} {Patterns} of {People} for
  {Compliant} {Robot} {Motion}},'' \emph{\BIBforeignlanguage{en}{The
  International Journal of Robotics Research}}, vol.~24, no.~1, pp. 31--48,
  Jan. 2005.

\bibitem{aoude_probabilistically_2013}
G.~S. Aoude \emph{et~al.}, ``\BIBforeignlanguage{en}{Probabilistically safe
  motion planning to avoid dynamic obstacles with uncertain motion patterns},''
  \emph{\BIBforeignlanguage{en}{Autonomous Robots}}, vol.~35, no.~1, pp.
  51--76, Jul. 2013.

\bibitem{trautman_unfreezing_2010}
P.~Trautman and A.~Krause, ``Unfreezing the robot: {Navigation} in dense,
  interacting crowds,'' in \emph{2010 {IEEE}/{RSJ} {International} {Conference}
  on {Intelligent} {Robots} and {Systems}}, Oct. 2010, pp. 797--803.

\bibitem{chen_decentralized_2016}
Y.~F. Chen, M.~Liu, M.~Everett, and J.~P. How, ``Decentralized
  {Non}-communicating {Multiagent} {Collision} {Avoidance} with {Deep}
  {Reinforcement} {Learning},'' \emph{arXiv:1609.07845 [cs]}, Sep. 2016, arXiv:
  1609.07845.

\bibitem{chen_socially_2017}
Y.~F. Chen, M.~Everett, M.~Liu, and J.~P. How, ``Socially {Aware} {Motion}
  {Planning} with {Deep} {Reinforcement} {Learning},'' \emph{arXiv:1703.08862
  [cs]}, Mar. 2017, arXiv: 1703.08862.

\bibitem{long_towards_2017}
P.~Long \emph{et~al.}, ``Towards {Optimally} {Decentralized} {Multi}-{Robot}
  {Collision} {Avoidance} via {Deep} {Reinforcement} {Learning},''
  \emph{arXiv:1709.10082 [cs]}, Sep. 2017, arXiv: 1709.10082.

\bibitem{everett_motion_2018}
M.~Everett, Y.~F. Chen, and J.~P. How, ``Motion {Planning} {Among} {Dynamic},
  {Decision}-{Making} {Agents} with {Deep} {Reinforcement} {Learning},''
  \emph{arXiv:1805.01956 [cs]}, May 2018, arXiv: 1805.01956.

\bibitem{helbing_social_1995-1}
D.~Helbing and P.~Molnár, ``Social force model for pedestrian dynamics,''
  \emph{Physical Review E}, vol.~51, no.~5, pp. 4282--4286, May 1995.

\bibitem{robicquet2016learning}
A.~Robicquet, A.~Sadeghian, A.~Alahi, and S.~Savarese, ``Learning social
  etiquette: Human trajectory understanding in crowded scenes,'' in
  \emph{European conference on computer vision}.\hskip 1em plus 0.5em minus
  0.4em\relax Springer, 2016, pp. 549--565.

\bibitem{kretz_indications_2018}
T.~Kretz, J.~Lohmiller, and P.~Sukennik, ``\BIBforeignlanguage{en}{Some
  {Indications} on {How} to {Calibrate} the {Social} {Force} {Model} of
  {Pedestrian} {Dynamics}},'' \emph{\BIBforeignlanguage{en}{Transportation
  Research Record}}, p. 0361198118786641, Jul. 2018.

\bibitem{sud_real-time_2007}
A.~Sud \emph{et~al.}, ``Real-time {Navigation} of {Independent} {Agents}
  {Using} {Adaptive} {Roadmaps},'' in \emph{Proceedings of the 2007 {ACM}
  {Symposium} on {Virtual} {Reality} {Software} and {Technology}}, ser. {VRST}
  '07.\hskip 1em plus 0.5em minus 0.4em\relax New York, NY, USA: ACM, 2007, pp.
  99--106.

\bibitem{ferrer_robot_2013}
G.~Ferrer, A.~Garrell, and A.~Sanfeliu, ``Robot companion: {A} social-force
  based approach with human awareness-navigation in crowded environments,'' in
  \emph{2013 {IEEE}/{RSJ} {International} {Conference} on {Intelligent}
  {Robots} and {Systems}}, Nov. 2013, pp. 1688--1694.

\bibitem{ferrer_robot_2017}
G.~Ferrer, A.~G. Zulueta, F.~H. Cotarelo, and A.~Sanfeliu,
  ``\BIBforeignlanguage{en}{Robot social-aware navigation framework to
  accompany people walking side-by-side},''
  \emph{\BIBforeignlanguage{en}{Autonomous Robots}}, vol.~41, no.~4, pp.
  775--793, Apr. 2017.

\bibitem{trautman_robot_2013}
P.~Trautman, J.~Ma, R.~M. Murray, and A.~Krause, ``Robot navigation in dense
  human crowds: the case for cooperation,'' in \emph{2013 {IEEE}
  {International} {Conference} on {Robotics} and {Automation}}, May 2013, pp.
  2153--2160.

\bibitem{trautman_sparse_2017}
P.~Trautman, ``Sparse interacting {Gaussian} processes: {Efficiency} and
  optimality theorems of autonomous crowd navigation,'' in \emph{2017 {IEEE}
  56th {Annual} {Conference} on {Decision} and {Control} ({CDC})}, Dec. 2017,
  pp. 327--334.

\bibitem{tai_socially_2017}
L.~Tai, J.~Zhang, M.~Liu, and W.~Burgard, ``Socially {Compliant} {Navigation}
  through {Raw} {Depth} {Inputs} with {Generative} {Adversarial} {Imitation}
  {Learning},'' \emph{arXiv:1710.02543 [cs]}, Oct. 2017, arXiv: 1710.02543.

\bibitem{long_deep-learned_2017}
P.~Long, W.~Liu, and J.~Pan, ``Deep-{Learned} {Collision} {Avoidance} {Policy}
  for {Distributed} {Multiagent} {Navigation},'' \emph{IEEE Robotics and
  Automation Letters}, vol.~2, no.~2, pp. 656--663, Apr. 2017.

\bibitem{liu2018map}
Y.~Liu, A.~Xu, and Z.~Chen, ``Map-based deep imitation learning for obstacle
  avoidance,'' in \emph{2018 IEEE/RSJ International Conference on Intelligent
  Robots and Systems (IROS)}.\hskip 1em plus 0.5em minus 0.4em\relax IEEE,
  2018, pp. 8644--8649.

\bibitem{pfeiffer_predicting_2016}
M.~Pfeiffer \emph{et~al.}, ``Predicting actions to act predictably:
  {Cooperative} partial motion planning with maximum entropy models,'' in
  \emph{2016 {IEEE}/{RSJ} {International} {Conference} on {Intelligent}
  {Robots} and {Systems} ({IROS})}, Oct. 2016, pp. 2096--2101.

\bibitem{mnih_human-level_2015}
V.~Mnih \emph{et~al.}, ``\BIBforeignlanguage{en}{Human-level control through
  deep reinforcement learning},'' \emph{\BIBforeignlanguage{en}{Nature}}, vol.
  518, no. 7540, pp. 529--533, Feb. 2015.

\bibitem{tai_virtual--real_2017}
L.~Tai, G.~Paolo, and M.~Liu, ``Virtual-to-real deep reinforcement learning:
  {Continuous} control of mobile robots for mapless navigation,'' in \emph{2017
  {IEEE}/{RSJ} {International} {Conference} on {Intelligent} {Robots} and
  {Systems} ({IROS})}, Sep. 2017, pp. 31--36.

\bibitem{alahi2017learning}
A.~Alahi \emph{et~al.}, ``Learning to predict human behavior in crowded
  scenes,'' in \emph{Group and Crowd Behavior for Computer Vision}.\hskip 1em
  plus 0.5em minus 0.4em\relax Elsevier, 2017, pp. 183--207.

\bibitem{antonini2006discrete}
G.~Antonini, M.~Bierlaire, and M.~Weber, ``Discrete choice models of pedestrian
  walking behavior,'' \emph{Transportation Research Part B: Methodological},
  vol.~40, no.~8, pp. 667--687, 2006.

\bibitem{sifringer2018let}
B.~Sifringer, V.~Lurkin, and A.~Alahi, ``Let me not lie: Learning multinomial
  logit,'' \emph{arXiv preprint arXiv:1812.09747}, 2018.

\bibitem{fernando_soft_2017}
T.~Fernando, S.~Denman, S.~Sridharan, and C.~Fookes, ``Soft + {Hardwired}
  {Attention}: {An} {LSTM} {Framework} for {Human} {Trajectory} {Prediction}
  and {Abnormal} {Event} {Detection},'' \emph{arXiv:1702.05552 [cs]}, Feb.
  2017, arXiv: 1702.05552.

\bibitem{sadeghian_sophie:_2018}
A.~Sadeghian \emph{et~al.}, ``{SoPhie}: {An} {Attentive} {GAN} for {Predicting}
  {Paths} {Compliant} to {Social} and {Physical} {Constraints},''
  \emph{arXiv:1806.01482 [cs]}, Jun. 2018, arXiv: 1806.01482.

\bibitem{liu2019collaborative}
Y.~Liu, P.~Kothari, and A.~Alahi, ``Collaborative gan sampling,'' \emph{arXiv
  preprint arXiv:1902.00813}, 2019.

\bibitem{sadeghian2018car}
A.~Sadeghian \emph{et~al.}, ``Car-net: Clairvoyant attentive recurrent
  network,'' in \emph{Proceedings of the European Conference on Computer Vision
  (ECCV)}, 2018, pp. 151--167.

\bibitem{liu_learning_2016}
Y.~Liu, C.~Sun, L.~Lin, and X.~Wang, ``Learning {Natural} {Language}
  {Inference} using {Bidirectional} {LSTM} model and {Inner}-{Attention},''
  \emph{arXiv:1605.09090 [cs]}, May 2016, arXiv: 1605.09090.

\bibitem{vaswani_attention_2017}
A.~Vaswani \emph{et~al.}, ``Attention {Is} {All} {You} {Need},''
  \emph{arXiv:1706.03762 [cs]}, Jun. 2017, arXiv: 1706.03762.

\bibitem{lin_structured_2017}
Z.~Lin \emph{et~al.}, ``A {Structured} {Self}-attentive {Sentence}
  {Embedding},'' \emph{arXiv:1703.03130 [cs]}, Mar. 2017, arXiv: 1703.03130.

\bibitem{hoshen_vain:_2017}
Y.~Hoshen, ``{VAIN}: {Attentional} {Multi}-agent {Predictive} {Modeling},'' in
  \emph{Advances in {Neural} {Information} {Processing} {Systems} 30}, I.~Guyon
  \emph{et~al.}, Eds.\hskip 1em plus 0.5em minus 0.4em\relax Curran Associates,
  Inc., 2017, pp. 2701--2711.

\bibitem{zhang_self-attention_2018}
H.~Zhang, I.~Goodfellow, D.~Metaxas, and A.~Odena, ``Self-{Attention}
  {Generative} {Adversarial} {Networks},'' \emph{arXiv:1805.08318 [cs, stat]},
  May 2018, arXiv: 1805.08318.

\bibitem{hochreiter_long_1997}
S.~Hochreiter and J.~Schmidhuber, ``Long {Short}-{Term} {Memory},''
  \emph{Neural Computation}, vol.~9, no.~8, pp. 1735--1780, Nov. 1997.

\bibitem{conneau_supervised_2017}
A.~Conneau \emph{et~al.}, ``Supervised {Learning} of {Universal} {Sentence}
  {Representations} from {Natural} {Language} {Inference} {Data},''
  \emph{arXiv:1705.02364 [cs]}, May 2017, arXiv: 1705.02364.

\bibitem{paszke_automatic_2017}
A.~Paszke \emph{et~al.}, ``Automatic differentiation in {PyTorch},'' Oct. 2017.

\bibitem{kingma_adam:_2014}
D.~P. Kingma and J.~Ba, ``Adam: {A} {Method} for {Stochastic} {Optimization},''
  \emph{arXiv:1412.6980 [cs]}, Dec. 2014, arXiv: 1412.6980.

\end{thebibliography}



\end{document}